\def\year{2021}\relax
\documentclass[letterpaper]{article} % DO NOT CHANGE THIS
\usepackage{aaai21}  % DO NOT CHANGE THIS
\usepackage{times}  % DO NOT CHANGE THIS
\usepackage{helvet} % DO NOT CHANGE THIS
\usepackage{courier}  % DO NOT CHANGE THIS
\usepackage[hyphens]{url}  % DO NOT CHANGE THIS
\usepackage{graphicx} % DO NOT CHANGE THIS
\def\UrlFont{\rm}  % DO NOT CHANGE THIS
\usepackage{natbib}  % DO NOT CHANGE THIS AND DO NOT ADD ANY OPTIONS TO IT
\usepackage{caption} % DO NOT CHANGE THIS AND DO NOT ADD ANY OPTIONS TO IT
\usepackage[toc,page]{appendix}
\usepackage{booktabs}
\usepackage{multirow}
\usepackage{color, colortbl}
\definecolor{airforceblue}{rgb}{0.36, 0.54, 0.66}
\definecolor{LightCyan}{rgb}{0.88,1,1}
\definecolor{codegreen}{rgb}{0,0.6,0}
\usepackage{array,etoolbox}
\preto\tabular{\setcounter{magicrownumbers}{0}}
\newcounter{magicrownumbers}
\newcommand\rownumber{\stepcounter{magicrownumbers}\arabic{magicrownumbers}}

\newif\ifpaperfinal

\paperfinaltrue

\ifpaperfinal
\newcommand{\tn}[1]{}
\newcommand{\iinote}[1]{}
\newcommand{\ernote}[1]{}
\newcommand{\stnote}[1]{}
\newcommand{\bknote}[1]{}
\else
\newcommand{\iinote}[1]{\textcolor{codegreen}{\textbf{II: #1}}}
\newcommand{\stnote}[1]{\textcolor{blue}{\textbf{ST: #1}}}
\newcommand{\tn}[1]{\textcolor{airforceblue}{\textbf{TN: #1}}}
\newcommand{\ernote}[1]{\textcolor{red}{\textbf{ER: #1}}}
\newcommand{\bknote}[1]{\textcolor{red}{\textbf{BK: #1}}}
\fi

\title{Where were my keys?  - Aggregating Spatial-Temporal Instances of Objects for Efficient Retrieval over Long Periods of Time}
\author{
    Ifrah Idrees, Zahid Hasan, Steven P. Reiss, and Stefanie Tellex\\

}
\affiliations{
    Dept. of Computer Science, Brown University, Providence, RI\\
}
\begin{document}

\maketitle

\begin{abstract}
Robots equipped with situational awareness can help humans efficiently find their lost objects by leveraging spatial and temporal structure. Existing approaches to video and image retrieval do not take into account the unique constraints imposed by a moving camera with partial view of the environment. We present a \textbf{D}etection-based \textbf{3}-level hierarchical \textbf{A}ssociation approach, \textbf{D3A}, to create an efficient query-able spatial-temporal representation of unique object instances in an environment. D3A performs online incremental and hierarchical learning to identify keyframes that best represent the unique objects in the environment. These keyframes are learned based on both spatial and temporal features and once identified their corresponding spatial-temporal information is organized in a key-value database. D3A allows for a variety of query patterns such as querying for objects with/without the following: 1) specific attributes, 2) spatial relationships with other objects, and 3) time slices. For a given set of 150 queries, D3A  returns a small set of candidate keyframes (which occupy only 0.17\% of the total sensory data) with 81.98\% mean  accuracy in 11.7 ms. This is  47x faster and 33\% more accurate than a baseline that naively stores the object matches (detections) in the database without associating spatial-temporal information.
\end{abstract}

\section{INTRODUCTION}
\iinote{color package in table}
\iinote{reduce font size of all tables}
\iinote{fix table 2}
\iinote{remove vspace}
Home-service robots have great potential to assist human users by retrieving spatial-temporal information\footnote{Spatial-temporal information of object refers to the whereabouts of the object such as where the object has been identified in the physical environment of the robot and at what times} about objects from their long-term observations. For example, a person can ask a simple query such as \textit{``where did I leave my keys?''} Service robots with such an ability will be well suited to help the elderly, especially those that have dementia.

There are two lines of works in image and video object retrieval: a) \textit{fixed-view cameras} \cite{object_retrieval, yadav2019vidcep, kang2019challenges} which assume full observability of the environment and do not deal with a moving camera (robot), and b) \textit{robotic object retrieval} \cite{ambrucs2014meta, bore2015retrieval} that either work on short-term time horizons (do not condense partial view detections of unique object instances for memory or speed efficiency) or perform long-term robotic object retrieval by specifically focusing on point cloud matching without incorporating object's spatial information in the physical environment. \iinote{object's spatial information in the physical environment, of where the objects are located in the environment}These approaches for object retrieval will leave the robot searching over countless detections in visual sensor data from many different time slices, which will take up a lot of space and time. In addition, these detections will contain partial views of different object instances, not all of these objects and their partial views will be\iinote{these objects and thus partial views will be} relevant to the query. An algorithm that condenses the partial view detections of each unique object instance into a compact and query-able spatial-temporal representation will enable the robot to answer queries about unique object instances more efficiently. 
\begin{figure}
    \centering
    % \includegraphics[width=0.8\linewidth]{images/story_2.png}
    % [width=0.8\linewidth,height=6cm
    \includegraphics[width=0.8\linewidth]{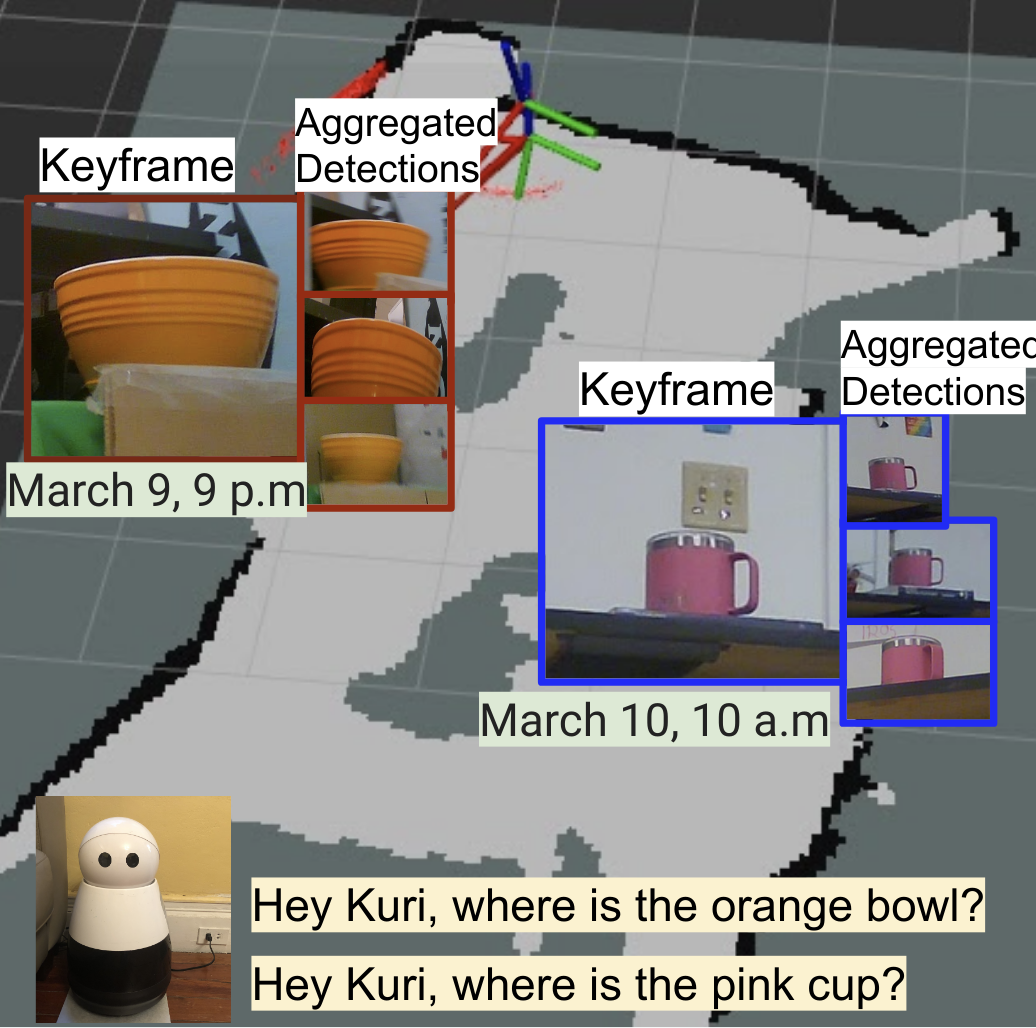}
    % \caption{Kuri Robot \citep{richardson_2018}  \stnote{I would replace this with a storyboard of the system, showing it answering a question about where an object is.  Also, make sure this is our picture of Kuri, use one you took that we have rights to.}}\label{fig:a}
    \caption{The image shows the retrieved keyframes and associated temporal information for two sample queries answered by D3A while deployed on Kuri robot. The keyframes are overlaid on the map using the spatial information returned by our spatial-temporal representation. Also shown are the partial view detections that are aggregated into the keyframe cluster.}
    \label{fig:Kuri}
\vspace{-4mm}
\end{figure}
To our knowledge, there has not been a single work that addresses all of the following challenges: 1) aggregating detections of object instances from different views in a map over hours of sensory information, 2) handling the object going out of the view, and 3) accounting for uncertainty in object poses due to noise from the robot pose. To mitigate these challenges, we introduce a new algorithm, D3A that extends the concept of keyframe extraction that has been previously used for video frame retrieval to condense
% \iinote{Add: that extends the concept of keyframe extraction that has been previously used for video frame(s) retrieval to condense}
multiview detections of unique objects over both physical space and time for long observation periods.

D3A extracts both spatial-temporal features of object instances in the environment as they get seen by the robot.
% Treating all these partial views of an object as unique and/or storing all of their spatial-temporal information will be memory and speed inefficient for object retrieval. To mitigate these challenges, 
D3A associates these partial view features \iinote{partial view features instead of views} with what it has previously seen over the map to condense them. 
% D3A enables scalable association and object retrieval by 
It does so by performing three-tier online, incremental clustering and filtering of unique spatial-temporal instances. While performing association, D3A keeps track of the keyframe that best represents each unique spatial-temporal location of the object. The information about keyframe-centroid clusters are then spatial-temporally indexed in a key-value database, leading to a compact and query-able representation for efficient object retrieval.

This representation allows D3A to identify and return a small set of keyframes for the object(s) in a given query, which enables a person to quickly find the relevant information. D3A is storage efficient and is able to handle the case where the attributes of object instances (color, shape, patterns) asked for in the queries are not known in advance. In prior work all or subset of these attributes were provided at the time of query.

Our algorithm is deployed on a social robot Kuri from  \citet{heykuri_2018} that patrolled a robotics lab\iinote{robotics lab} environment and collected 22 hours of observations over four days\footnote{\label{note1} A visualization recording of the initial state of the environment, along with its 2D map and sample of the collected dataset can be found at this link: https://github.com/IfrahIdrees/D3A.git}. When tested on a set of 150 queries about different objects, D3A returns a small set of keyframes and their spatial-temporal information (which \iinote{on average} occupy only 0.17\% of the total sensory data\iinote{sensory data of object of interests}) and is 47x faster than the baseline model which treats all partial-view detections as independent instances. D3A is also 33\% more accurate than the baseline with an accuracy value of 81.98\%.

\section{Related Work}
In the past, \textbf{Keyframe Extraction} techniques have been used to extract the summary key frames to represent video sequences \cite{6135507}.
% \cite{5201141, 6135507, lingam2019key, event, object_tracking}. 
A lot of different techniques exists for keyframe extraction such as  clustering \cite{vazquez2013spatio}, energy minimization \cite{essa2015modular}, or online techniques \cite{elahi2020online}.
% These differ in how they define the similarity measure between visual descriptors of the frames. 
% \tn{the previous sentence feels out of place and should be cut. You can just say that D3A extendes the clustering keyframe extraction technique to ...} 
D3A extends the clustering keyframe extraction technique in videos to incrementally find keyframes that best represent the different views of the unique objects in the environment over time, as the robot continues to capture new video frames.
% \iinote{Add: In the past \textbf{Key frame Extraction} techniques have been used to extract the summary key frames to represent video sequences \cite{5201141, 6135507, lingam2019key, event, object_tracking}. A lot of different techniques exists for keyframe extraction such as  clustering, energy minimization  or sequential techniques \cite{vazquez2013spatio}. These differ in how they define an appropriate similarity measure between visual descriptors of the frames. D3A extends keyframe extraction in videos to incrementally finding keyframes that best represent multi views of the unique objects in the environment over time, as the robot continues to capture new video frames.}

A lot of work has also been done in \textbf{Object Detection} \cite{MaskRCNN, RetinaNet, FastCNN, pose_estimation} for detecting and re-identifying objects. D3A is built upon the notion of object detection in a three-tier mechanism to match objects and condense their detections over long periods of observations. Additionally, there are \textbf{Object Tracking} algorithms that track object instances over time  \cite{object_tracking, yilmaz2006object}. However, they usually assume input videos with fixed angles and cannot track/associate objects under occlusion or after the objects go out of view.
% \iinote{In past keyframes is a method used to extract the perform clustering of consecutive frames to find representative keyframes for all the consecutive frames video trying to find unique views of scene. we extend that to actually find keyframes  to associate multi views of the object for object association over space and time.They differ in how they define an appropriate similarity measure between visual descriptors of the frames. }
% These algorithms don't scale well with the number of objects. 
These algorithms are also not useful for retrieval since they do not focus on making the objects' trajectories query-able.
% \tn{can cut: We use an unsupervised clustering method in our implementation which can be replaced by any other techniques.}
% \ernote{What other techniques are likely? I feel like you mean that you chose a specific unsupervised clustering algorithm X, but any unsupervised clustering algorithm could be used?}

Although many \textbf{Image or Video Retrieval} systems have been introduced in the multimedia community, they focus on retrieval at the level of object classes and not instances. These works can be categorized into the following types:\\
\textit{1) Fixed-angle Video Input} \cite{object_retrieval, yadav2019vidcep, kang2019challenges}: These works do not account for the objects' spatial information in a global map. In contrast, D3A can perform object instance association in a partially observable environment monitored by a robot.\\
\textit{2) Query Optimized Object Retrieval} \cite{kang2019challenges}: This work uses information from the query to optimize the retrieval pipeline by searching over all the scaled down image frames in the original dataset. They do not reduce redundancy in the data,
%  Their search space is over all the frames and they do not focus on reducing the redundancy in data \cite{kang2019challenges}. 
%  Our algorithm instead focuses on optimizing the spatial-temporal representation to be used for retrieval. We focus on condensing the detections to only store key frame information for retrieval. 
whereas D3A focuses on only storing the keyframe information 
% to condense the detections 
and optimizing the  spatial-temporal representation.

There are also works that can handle partial views of objects and perform view-invariant object detection at the category level \cite{hsieh2018focus, sivic2003video}. 
% However, they do not record the objects' spatial information and cannot answer spatial queries like D3A. 
\citet{hsieh2018focus} proposed work similar to ours that can handle many hours of videos captured by a moving camera (that is not a robot) but does not use object locations. We perform an ablation study with this work and show that by using the objects' locations in the map, D3A improves 1) detection of unique object instances and 2) compactness and retrieval performance from our spatial-temporal database. \citet{sivic2003video} assume to have seen a part of the scene captured by different static camera viewpoints to learn the visual vocabulary (which incorporates spatial information) to associate objects seen in later scenes. We perform online, incremental object matching, and do not pre-define the number of clusters/objects in the environment. 
% Unlike both of these works, our clustering is based on both the object feature space and spatial-temporal information. 

% There are other works that deal with robotic object retrieval but over short-time horizons and therefore they do not condense the partial views for memory and speed efficiency. 
% Another relevant line of work is robotic object retrieval. Most works in this line operate over short-time horizons and do not condense partial view detections of unique object instances for memory or speed efficiency \cite{Temporal-Grounding} \tn{feels like you're missing citations}. 
% \tn{what's the difference between these works and the ones in the previous paragraph?}
There are also works that specifically focus on long-term robotic object retrieval that handle partial observability in visual sensor data \citep{ambrucs2014meta, bore2015retrieval}. However, these works assume a point cloud of objects as input and introduce matching  algorithms specifically in 3D space. D3A, in contrast, does matching using the RGB feature space and limited depth information (information provided by the depth sensors for obstacle detection) that can be derived from the 3D space. \tn{is this better bc it requires ``less" visual information or just different} These works also do not deal with the objects' spatial information in the environment. Furthermore, they do not focus on making a compact representation of the sensory data for retrieval, but instead store and operate over the original data and use the object attribute ``shape" provided in query to optimize retrieval. 
% \tn{This also seems similar to the query optimized object retrieval works described above. Are these the works that require that the queried object attributes are known in advance? This bit is not explicitly stated anywhere}

% \stnote{use citep, not citet, only use cite for noun phrases.  for multiple cites, put them in one citep command separated by commas }

% \begin{figure}[t]
%     \centering
%     \includegraphics[width=\linewidth]{images/high_level.png}
%     \caption{Block Level diagram of D3A \tn{D3A}}
%     % \roma{Trim the edges so they fit exactly and also make the figure width the line width so this fits the page nicely and the font is easier to read :-)}}
%     \label{fig:a}
%     \label{fig:block-diagram}
% \end{figure}

% \begin{figure}[t]
%     \centering
%     \includegraphics[width=\linewidth]{images/1.jpg}
%     \caption{\small Visualization of D3A Algorithm \iinote{Add Qi, si}}
%     % \label{fig:a}
%     \label{fig:Processing}
% \vspace{-6mm}
% \end{figure}

\begin{figure*}[t]
\centering
\includegraphics[width=0.75\textwidth]{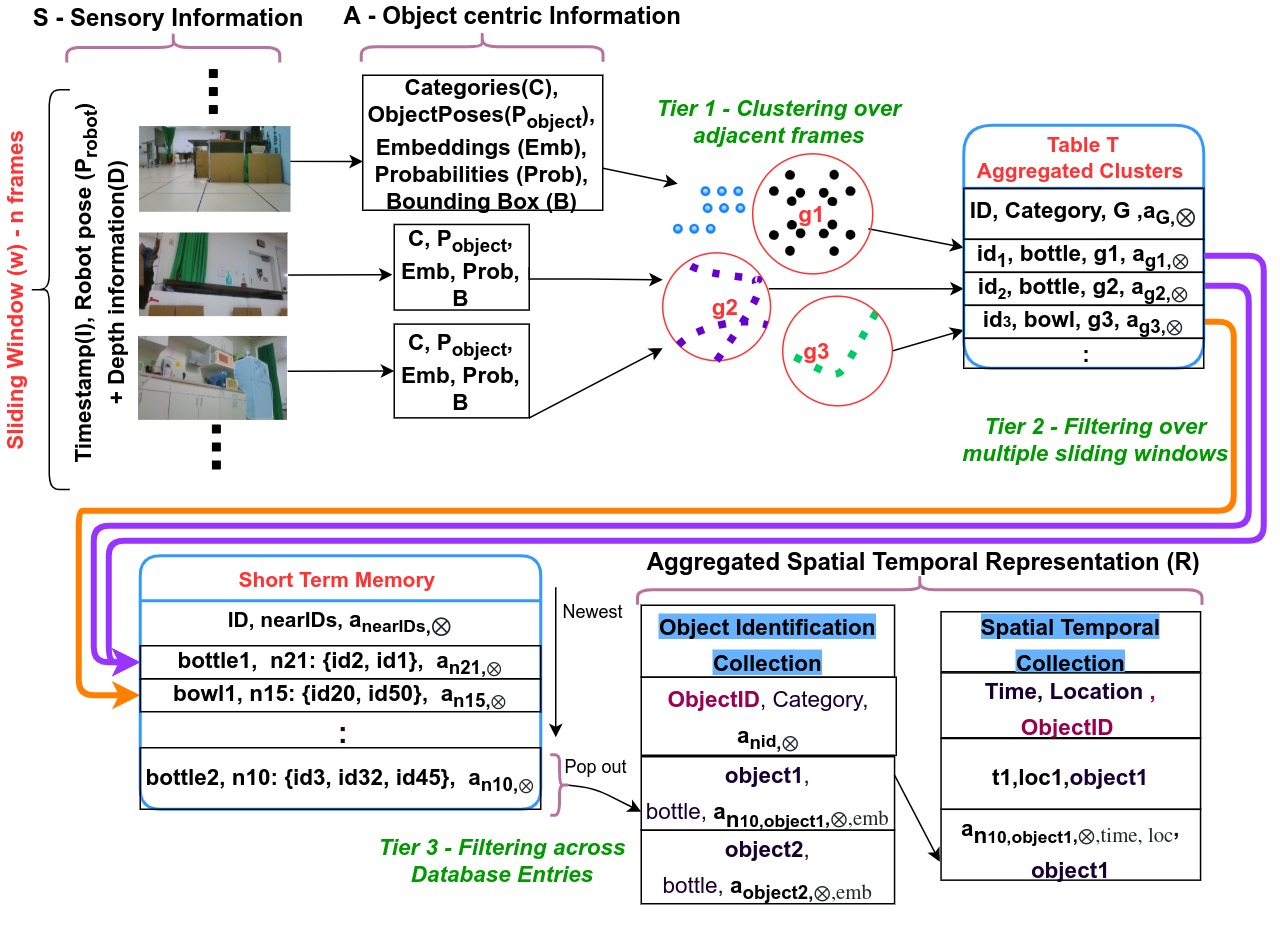}
\caption{\small Visualization of D3A Algorithm \iinote{Add Qi, si}}
\label{fig:Processing}
\vspace{-3mm}
\end{figure*}

%  \tn{this is super readable now! i recommend using a different color (darker green maybe) to label the tiers bc the current color is a little hard to see (same for Fig 3)}

\section{Problem Formulation}\label{Section:system_overview}
% \stnote{So you want to write technical approach in a more problem-driven way.  Think of it as a tree of problems and solutions.  You don't want to introduce a bit of technical material until you've first explained the problem you need to solve by that bit.  The paragraph below is just saying what you built without saying why or situating it in what problems you are solving.}
% \iinote{how link speed and memory efficiency}

\iinote{fix time variable as t?}

A robot monitoring the environment over long periods is given a spatial-temporal query $Q$ to retrieve spatial temporal information of target objects seen in the environment. The query can be of any type as listed in Table-\ref{tab:example_queries}. The robot is equipped with the map of the environment and at any given time step $i$ can gather the following sensory information $s_i=<p_{robot,i}, d_i, f_i>$, where $p_{robot,i} \in P_{robot}$ is the associated robot's pose, and $d_i \in D$ is the corresponding depth information for image frame $f_i$. The robot collects large amounts of sensor data $S = \{s_1, s_2, ... , s_i\}$ and uses an object detector to extract the relevant object-centric information $a_i$ from every frame $f_i$. However, the robot has partial observability of the environment and the information required to answer the object retrieval query is going to be dispersed throughout $S$ in the form of multiple partial view detections. To answer object-retrieval queries efficiently, the robot needs to condense the multi-view detections of the objects into a memory efficient and query-able representation. This representation is then used by the robot to return a small set of keyframes, which enables a person to quickly find the  relevant information about the target object. 
% The sensor data $S$ will take up a lot of disc space and be inherently noisy. Searching all of this data for retrieving object's spatial-temporal information is  going to be speed-inefficient. 
% % The robot should also be able to answer queries where the attributes of object instances (color, shape, patterns) are not known in advance. 
% For the robot to retrieve an object's spatial temporal information in a memory and speed efficient, it should be able to condense the multi-view detections of the objects over time and be able to organize them in queryable representation such that given a query $Q$, it can return a small set of keyframes, which enable a person to quickly find the  relevant information about the target object. 

\section{Approach}
% all this sensor data $S$ through which the object need to be retrieved takes up a lot of disc space and is inherently noisy.

% Equipped with $S$, the robot is given a spatial-temporal query $Q$ of any of the type listed in Table-\ref{tab:example_queries}. One example is ``Did  you  ever  see  an  orange bowl" and is tasked to retrieve spatial temporal information of target objects seen in the environment. Ho 

% A human given a robot collecting $S$ can ask from their personal assistant robot a spatial-temporal query - $Q$ about any object in the environment seen during any time of its patrol such as can be  "Did  you  ever  see  an  orange bowl". 

% $=< p_{robot,i}, d_i, f_i>$ is the associated information for every captured image frame $f_i$. $p_{robot,i} \in P_{robot}$ is the associated robot's pose, and $d_i \in D$ is the corresponding depth information, for every image frame $f_i$. 
% Note that the next time step $i+1$ is dependant on the frequency at which the robot is capturing the sensor data.
% All this sensor data $S$ takes up a lot of disc space and is inherently noisy. 
For efficient spatial-temporal object retrieval in a partially observable environment, we need to compress this data $S$ into a compact and query-able spatial-temporal representation $R$. For this we first pre-process every $s_i$ to extract the relevant object-centric information $a_i$, then apply a three-tier online clustering algorithm to aggregate $a_{i}$'s. The output of D3A is a set of clusters of unique object instances that can be aggregated into the previous time step's compact representation $R_{i-1}$ to create a new representation $R_i$ of the environment for time step $i$. A diagram of our algorithm creating $R$ is shown in Fig-\ref{fig:Processing}.  Details of algorithmic implementation can be found in the Appendix.
% \stnote{problem formulation is only the problem, not the solution.  So don't put anything about preprocessing etc in the problem formulation.}
% \vspace
\subsection{Pre-processing each frame:} 
% For compactness and speed efficiency, D3A needs to reduce the dimensionality of the visual sensory data $f_i$ in $s_i=< p_{robot, i}, d_i, f_i>$, as well as aggregate multiple detections of the same object over long periods of time. D3A achieves this by extracting minimal amounts of relevant object-centric information $a_i = <C_i, Prob_i , B_i, P_{object,i}, Emb_i>$ for every frame $f_i$. There is a set $J_i$ for every $f_i$ that includes all the detected object instances in $f_i$. $C_i$ is the list of corresponding categories for all the detections in the set $J_{i}$,
% % for each detection $j \in J_i$ and $J_i$ represents the set of all detections in $f_i$, 
% $Prob_i$ are the probabilities of detecting the object categories $C_i$, $B_i$ is the list of the object bounding boxes for $J_i$, $P_{object,i}$ is the list of object poses estimated for all the detections. $P_{object,i}$ is estimated from the robot's pose $p_{robot,i}$ and limited depth information $d_i$, and is a 3D world coordinate. $Emb_i$ is the list of embeddings corresponding to the detections in $J_i$, which is used to relate $j's$ over time and create the compact representation $Q$. 

For compactness and speed efficiency, D3A needs to reduce the dimensionality of the visual sensory data $f_i$ in $s_i=< p_{robot, i}, d_i, f_i>$, as well as aggregate multiple detections of the same object over long periods of time. D3A achieves this by extracting minimal amounts of relevant object-centric information ${a_i}$:
\begin{center}
$a_i = [ <c_{i1},prob_{i1},b_{i1},p_{robot,i1},emb_{i1}>, <c_{i2},prob_{i2} ,b_{i2},p_{robot,i2},emb_{i2}>,..., <c_{ij}, prob_{ij},b_{ij},p_{robot,ij},emb_{ij}>]$
\end{center}
for every frame $f_i$, $c_{ij} \in C_i$, is the category detected for the $j^{th}$ object instance detection in $i^{th}$ image, $prob_{ij}$ is the probability of $j^{th}$ object being assigned the category $c_{ij}$, and $b_{ij}$ and $emb_{ij}$ is the bounding box and object embedding for the $j^{th}$ detection in $f_i$. $emb_{ij}$ is used to relate it with the other objects over time and create the compact representation $R$.

\subsection{Tier 1- Clustering across adjacent frames}

% \tn{increase font size for text outside boxes}
% \stnote{This paragraph is a really nice example of introducing the problem first.}
The extracted meta-information for the $jth$ detection in the $ith$ frame -- $a_{ij}$ is noisy because of the uncertainty in the robot's pose estimate $p_{robot,i}$ and noise in the extracted object-instance centric information -- $c_{ij}$, $prob_{ij}$, $emb_{ij}$. To aggregate the different view detections of the same object and to identify the unique objects in the environment, we consider a sliding window $w$ over sensory data $\{s_{i - n}, ..., s_i \} $. Object-centric information $a_{ij}$ of all detections in all the frames of this sliding window are then clustered to create a set of clusters $G$, each identifying a unique object instance. The motivation behind this is that the object instances of same category with similar embeddings and positions in space will be associated to the same cluster hence pruning the noisy detections. We assign a unique id $instance_{id} \in ID$ to each cluster $g \in G$ and perform an aggregation operation $\bigotimes$ on every cluster $g \in G$ to create a unified feature representation of each of the cluster. This results in an aggregated representation $a_{g,\bigotimes}$ which is then inserted into a collection $T$ indexed by the unique id $instance_{id}$ of each cluster where $ID$ is the set of all $instance_{id}$'s \iinote{fix this}.

\subsection{Tier 2 - Filtering across multiple sliding windows}
% \begin{figure}[t]
%     \centering
%     \includegraphics[width=0.6\linewidth]{images/Database_design.png}
%     \vspace{-2em}
%     \caption{Schematic Design of Database . *2 -Set of time intervals tuples containing start and end time.
% *3.- Associated with each time interval is a bounding box and  key frame \tn{the font's a little small, change the green color to be darker, and what is PK \& FK?}}\label{fig:a}
%     \label{fig:Schema}
% \end{figure}
% \stnote{introduce problem first}
We also need to aggregate the detections of unique instances across the objects' multiple partial views in the past. This requires sharing information across multiple sliding windows via a second-tier filtering. To facilitate this, we maintain a fixed-length short-term memory (STM) indexed over $ID$ \bknote{read this?}and aggregate information in it for all instance id's in $T$. 
% The ouput of Tier 1 ($ID$ in Table $T$) still needs to pass through more filtering since the detections of unique object instances need to be aggregated with the previous objects' multiple partial views for further compactness and association. This requires sharing information across multiple sliding windows via a second-tier filtering. To facilitate this, we maintain a fixed-length short-term memory $(STM)$ that is indexed over $ID$. 
For an object represented by $instance_{id}$ in collection $T$,
we add a new entry to STM only if the object is a new unique object. Otherwise, if the object has previously been seen by the robot, an entry for this object should already exist in STM and should be updated.
We do this by finding a set of $nearIDs$ in STM that are closest in embedding space in cosine similarity\iinote{tell here?} and measuring the euclidean distance between them and $instance_{id}$ in physical space. There are three cases to be considered here that symbolize whether the object is new or not and whether it has moved or not: \textbf{1)} If $nearIDs$ set is non-empty and euclidean distance is less than a threshold $d_{thresh}$ then we perform another aggregation operation $\bigodot$ on $instance_{id}$ and the closest $n_{id} \in nearIDs$ to create a unified feature representation for $n_{id}$. This results in an aggregated representation $a_{n_{id},\bigodot}$ that is updated in $STM$. \textbf{2)} If $nearIDs$ set is non-empty and euclidean distance is greater than a threshold $d_{thresh}$, we make a new entry in the STM with instance id $instance_{id}$. This new entry with the same $instance_{id}$ symbolizes dynamic objects (where the same object has moved in the environment). \textbf{3)} If $nearIDs$ is empty for $instance_{id}$, we also add a new entry in the STM with instance id $instance_{id}$. This represents the scenario where the object has not been seen before.
After the aggregation operation, the keyframe of $n_{id}$ is chosen to be the frame with the highest detection probability $prob_{ij}$ among the frames $f_i's$ associate with $instance_{id}$ and $n_{id}$.  If a new instance is to be added to the STM when it has reached its maximum capacity, our algorithm evicts the last recently viewed (least-recently viewed entry), $lrv\_entry$, from the STM and moves it to the persistent storage key-value store $R$. This eviction strategy ensures that objects currently being viewed by the robot remain in the STM for further noise filtering and aggregation of meta-data from partial views.

% \iinote{we do it for all the $instance_{id}'s$ in the window.} \iinote{key-value database}
\subsection{Tier 3- Filtering across Database Entries} \label{subsubsection:p3}
The filtered entry $lrv\_entry$ from the STM could be directly added as the final aggregated cluster into our compact representation $R$. However, we want to aggregate detections of unique object instances over not just recent times but also long periods of time.
% However, we want a more compact representation that aggregate detections of unique object instances over long periods of time.
To do so, every time a $lrv\_entry$ is evicted from STM, our third-tier filtering level looks up and updates the aggregated clusters stored in $R$. We organize our aggregated spatial-temporal representation $R$ in a key-value database over two collections. The \textbf{Object Identification Collection} ($OIc$) is updated using the object information of the evicted clusters represented by the $lrv\_entry$, while the other store \textbf{Spatial-temporal Collection} ($STc$) stores the object entries indexed over time and position in physical space. Every object inserted in $OIc$ is associated with a unique identifier $ObjectID$ which is used to index $STc$ to get the object detections over space and time.

To update $R$, we find the set $Records \in OIc$ that are the closest in feature space with $lrv\_entry$ in cosine similarity. A new document is inserted into $STc$ if the normalized Euclidean distance between last evicted entry from STM $lrv\_entry$ and $Records \in STc$ is greater than $d_{thresh}$, otherwise the object information in the closest document $record \in Records$ is aggregated, via an operation $\oplus$, applied on the $lrv\_entry$ and the document $record$ hence updating the object information in $OIc$ using this aggregated output. This new document insertion in $STc$ symbolizes a moving object while the update symbolizes operation for a static object. For every new entry in $OIc$, an entry is added in the $STc$ against the $ObjectID$ containing the time and physical location of the new entry.
\subsection{Query Processing and Answering}
\begin{table}
\small
\centering

%\stnote{Can you add a column for the formal representation of the query?} \iinote{I had that previously but it was taking too much space so removed it? Will it be okay if we dont have it?}
\begin{tabular}{p{0.4cm}p{3cm}p{3.4cm}} 

\toprule
  \bf{Type} & \bf{Language Example} & \bf{Formal Representation} \\
\midrule
Q1 & Did you ever see an orange bowl? & ${orange(x), bowl(y) |}$ ${ OIc.find(x,y) }$\\
\hline
% Q2 & { "Where was the $attr1$ pink cup"?}  \\
% \hline
Q2 &  When were the cup and bowl seen together? & ${bowl(x), cup(y) |}$  ${STc.find(}$ ${OIc.find(x,y), together)}$\\
\hline
Q3 & Where did you see the pink cup between 1:00pm and 4:00pm on 1st Jan, 2020? & ${pink(x), cup(y), time(t) |}$ ${STc.find(OIc.find(x, y),}$ ${1600 >= t >= 1300)}$\\
% \hline
% Q4 & Where is the Sprite bottle \\

\bottomrule
% \label{example_queries}
\end{tabular}
\vspace{-2mm}
\caption{Query Types, Language Examples \& Formal Representations}
\label{tab:example_queries}
\vspace{-6mm}
\end{table}
Template-based NLP methods \cite{li2017querying} or deep learning \cite{DBLP:journals/corr/abs-1911-04942} methods\tn{probably need citation} can be used to map a language query to a set $X$ of $ObjectID's$, time intervals, and location areas, which D3A takes as input for object retrieval. Table \ref{tab:example_queries} shows the types of queries, their natural language examples, and formal representations used in our evaluation. This formal representation is used to search over indexed $OIc$ and $STc$ collections of $R$. The representation $R$ built through our method can handle a wide variety of queries, including: querying for objects with/without the following: 1) specific attributes such as ``orange bowl," 2) spatial relationships with other objects such as ``cup and bowl together," and 3) time slices which can be either a point in time or a time interval. Our representation can also efficiently handle negative queries where the answer does not exist in the dataset.

% objects with/without 1) specific attributes, 2) spatial relationship among objects, 3) the temporal relationship among objects, and 4) time slices. 

% The query input to RoboMem is a combination of $ObjectID's$, time intervals and location area. This set can be generated from Natural Language Query and sample object image. However, our current system takes in a templatized query. The schematic design of the database can handle a variety of queries as follows: Q1, Q2: With specific attributes (for static objects and dynamic objects), Q3) and Q4) With time slices(for positive and negative query). A negative query is the one for which the answer in the database does not exists. Q5: with a query about latest time stamp, Q6, Q7: with spatial relationship among objects.  Fig-\ref{fig:block-diagram} show examples of the query types in natural language and their corresponding DB query. 

% 3) the temporal relationship among objects, and 4) time slices.
% \begin{enumerate}
%     \item With specific attributes: "Where was the $attr1$ $catg$"? mapped to 
%     \begin{verbatim}[scale =0.99]
%     Q1: object\_id=collection.find({Category: attr1}, feature vector: attr1)
%         collection.find({obj\_id : object\_id})
%     \end{verbatim}
%     \item 2) spatial relationship among objects, 3) the temporal relationship among objects, and 4) time slices.
% \end{enumerate}

\section{Evaluation}
% \stnote{The aim of our evaluation is to test the hypothesis that algorithm X improves the precision and recall of a database-backed system and answering queries.}
\begin{table}[t]
    \small
    \centering
    \begin{tabular}{l p{0.5cm} r}
    \toprule
     Duration &22hrs & 3hrs \\
    \midrule
    Total Number of Frames & $10,132$ & $991$\\
    Frame Rate (per minute) & $7.67$ & $7.67$\\
    Sensor Data Size (GB) & 10.53 & $1.43$\\
    Total Number of Detections & $13,565$ & $2,558$ \\
    Ground Truth Objects & N/A & $59$ \\
     & & Static = 49, Dynamic = $10$\\
    Number of Times & N/A & Mean = $2.1 \pm 1.3$, \\
    Dynamic Objects move && Max = 5.0 \\
    \bottomrule
    \end{tabular}
    \vspace{-3mm}
    \caption{Dataset Details}
    \vspace{-3mm}
    \label{tab:my_label}
%\vspace{-3mm}
\end{table}

The aim of our evaluation is to test the hypothesis that a database-backed system with our algorithm D3A improves both 1)  the compactness of the spatial-temporal representation of objects in the environment and 2) the object retrieval performance by returning a small subset of keyframes as measured by the mean reciprocal rank, miss rate  and retrieval time as described in Experiment Design subsection. 
% Sec-\ref{sec:exp-des}. 
As a result, the user will have to search over just a few returned results to find the answer to their question.
To test our hypothesis, we perform a real-world evaluation, using both a range of synthetic detectors built from ground truth data to control the noise level and real detectors deployed on a mobile robot to assess the system's end-to-end accuracy. 

% This is directly related to improved user experience. 
%%%\stnote{So instead of telling that, can you like SHOW that?  As a result, the user will have to search few just a few returned results to find the answer to their question.}  \stnote{To test this hypothesis, we perform a real-world evaluation on a mobile robot using both simulated detectors built from ground truth data to test the noise  and real detectors to assess the end-to-end accuracy.  (PS, I am writing this fast, so feel free to reword/clean up any suggested sentence I put.  But it's good to have something to say like, what the evaluation is going ot be and why it will answer our hypothesis.) }

% Our main results are:
    % \beging{}
% and hence the recall\footnote{Defined as a function of true positives (tp: key frame containing the object of interest in the query), false negatives (fn: objects failed to be identified and retrieved). It is the fraction of relevant documents that are retrieved, in particular Recall = \frac{tp}{tp+fn}} 
% measure for evaluatingreturned key frames to a sample of queries, ordetred by probability of correctness. The reciprocal rank of a query response is the multiplicative inverse of the rank of the first correct answer. If none of the proposed results are correct, reciprocal rank is 0.} 
% with varying range of sensor uncertainty for retrieved objects. 
    
\subsection{Dataset Collection}
% \stnote{don't put colons after subsection names}

% We deployed D3A on a mobile robot Kuri \cite{heykuri_2018} that patrolled a robotics lab environment for four days. 
The robotics lab environment in which our mobile robot Kuri \cite{heykuri_2018} patrolled for four days was uncontrolled and cluttered: people were allowed to use the space as is. For consistency, we kept the illumination same throughout the data collection (it was captured in a completely lit lab). The robotics lab area used for data collection and experimentation included the kitchen area and the general area with tables and chairs. This real-world dataset\footnote{ A visualization recording of the initial state of the environment, along with its 2D map and sample raw sensory data can be found at \url{ https://github.com/IfrahIdrees/D3A.git}.}
has 10,132 image frames over 22 hours and contains both static and dynamic objects such as various cup and bottles changing there location.  The dataset details can be found in Table \ref{tab:my_label}. 
% \iinote{for consistency keep the illumination same(captured during daytime.)}
% \stnote{cut: To our knowledge this is currently the longest dataset collected by a robot.}  \stnote{No way.   For example see the MIT SLAM dataset.  And all the Waymo selfdriving data.  }

\subsection{Dataset/Object Annotation}
\begin{figure}[t]
    \centering
    \includegraphics[width=0.9\linewidth]{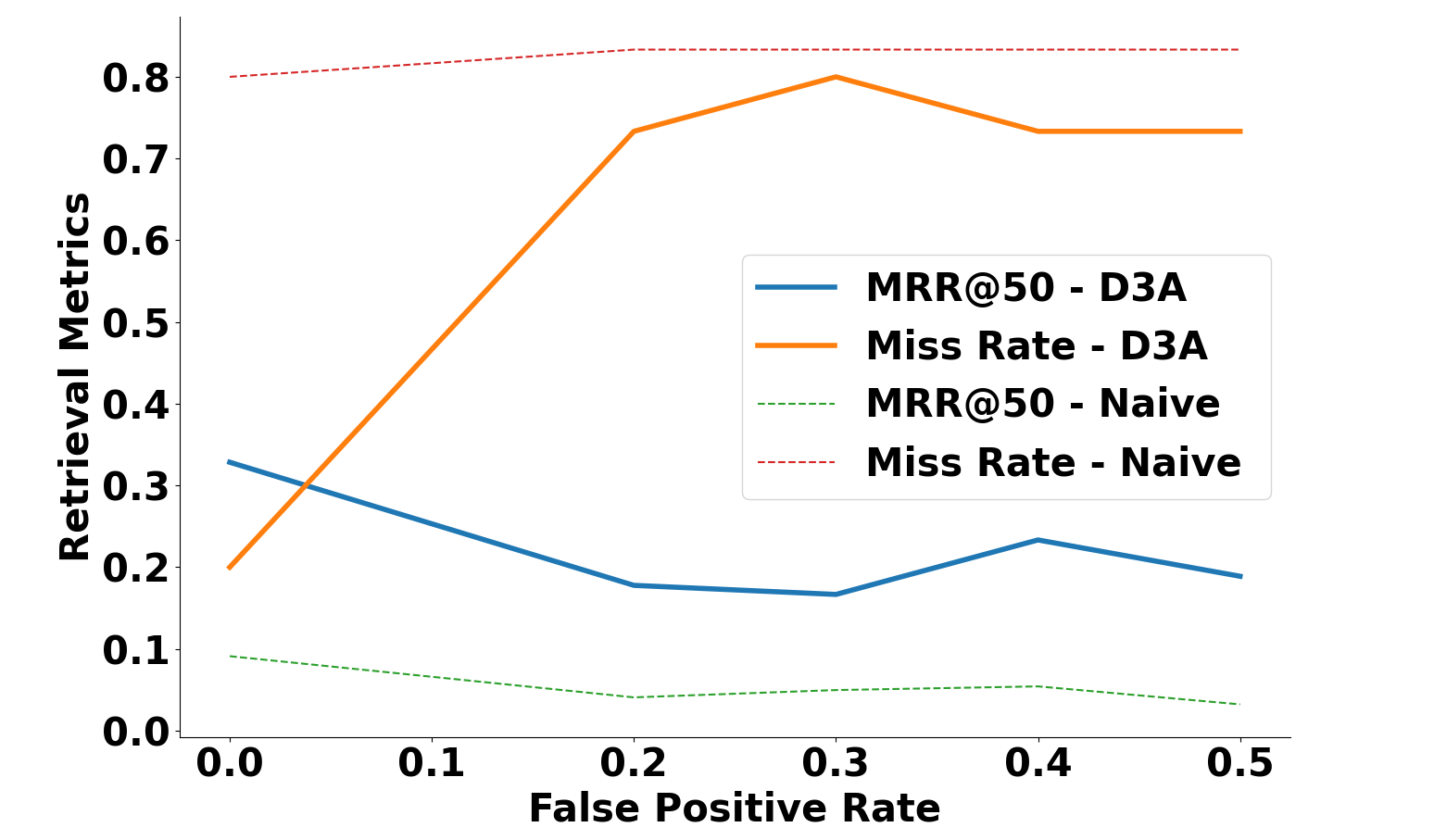}
    \caption{Exp 1 - Retrieval performance with a synthetic embedding generator with increasing uncertainty}
    % \label{fig:a}
    \label{fig:graph_noise}
\vspace{-4mm}
\end{figure}

% \stnote{No colons in section names.}
% \stnote{I take back the comment in the intro about 22 hours vs 3 hours, but we should make sure we use the 22 hours somehow in the evaluation. }
For evaluation purposes, the collected dataset was manually annotated by the authors. Every unique instance in the dataset was assigned a unique id and a ground truth location that was used to perform object association over time and space. Due to time and manual labour constraints, we were only able to annotate 3 hours of data.

\subsection{Parameter Selection}
% \stnote{No colons in section names.}
% \iinote{ We use a color histogram \cite{color_hist} to approximate an embedding of each object instance to relate them over time. % and reduce the set of potentially similar objects. We select the color histogram method due to its simplicity,}
The Euclidean distance threshold  $d_{thresh}$ , sliding window length, normalized threshold for feature matching (cosine similarity), and size of short-term memory (STM) were set to 0.5m, 10, 0.4, and 400 (our RAM's maximum capacity), respectively. We performed a parameter sweep offline on the collected data and found this setting to be optimal. Increasing the feature matching and distance threshold increased the number of false positives during tier-1 processing. For the STM, the greater the buffer size, the more noise there will be in the object's pose and embedding to be filtered and processed.

% The selected parameter values and the justification for each are as follows:

% 1) The Euclidean distance threshold $d_{thresh}$ was set to 0.5 meters to account for the 0.5m standard deviation in the robot pose estimation. We performed a parameter sweep and found that setting the sliding window length = 10 and normalized threshold for feature matching = 0.4 to be optimal. Increasing the matching and distance threshold increased the number of false positives during tier-1 processing.
% 2) The short-term memory (STM) buffer was set to 400 objects, our RAM's maximum capacity. The greater the buffer size, the more noise there will be in the object's pose and embedding to be filtered and processed.
% We also performed a parameter sweep over different lengths of the buffer and found the results to be not impacted

\subsection{Experiment Designs}\label{sec:exp-des}

We conduct four experiments with two baselines. One experiment is a running example comparison.
% We conduct three \tn{four?} sets of quantitative experiments with a running example and two baselines (as described below). 
% \stnote{Don't give results here until after you describe the algorithms.  So move those two sentences later.}

\textbf{Baselines}: Our first baseline is a ``Naive" method that inserts every detection's spatial-temporal information directly into the database without aggregation. This baseline ablates the three-tier processing that is integral to our system for aggregating partial views of the objects, but can still index detections on space and time. We also conduct an ablation of the spatial information from the features to make a ``Non-spatial" baseline, similar to \citet{hsieh2018focus} which does not utilize spatial information during clustering, to compare the clusters resulting from both methods. 
% We do another ablation study where we removing the spatial information used by D3A for three-tier clustering. We refer to this method as ``Non-spatial''. This is similar to \cite{hsieh2018focus} wich uses a similar clustering method but does not use the spatial information. 

% For an ablation study with \citet{hsieh2018focus}, we implement a ``Non-spatial'' baseline, removing the spatial information used by D3A for three-tier clustering. We use a running example to compare D3A with this baseline and visualize the differences in their aggregated clusters. 
%%%\stnote{I would rephrase that a bit, and say first, we do the abalation, and then, this is similar to hsieh whichuses a similar clusting method but does nto use spatial information}
% will show \tn{that clustering by spatial information improves...} how D3A improves the aggregation of detections.

% In the first, we show the changing trend in retrieval performance of our system relative to the uncertainity of the parameterized synthetic detector as compared to our baseline. 
% We perform two separate ablation studies that extensively assess the improvements from our framework.

\textbf{Metric Definitions}: 
We measure the quality of keyframes returned for a query with the MRR@50 evaluation metric, which is the multiplicative inverse of the rank of the correct keyframe in the top 50 returned frames ordered by their probability of correctness. We want the rank of the correct frame to be as close as the start of the list of returned frames and hence higher MRR closer to 1 is better. We measure the miss rate (the number of objects missed by the algorithm during the clustering phase due to detection or classification error, hence cannot be found in the returned frames) to get an idea of our accuracy. We also report the speed efficiency of our system by measuring the mean number of frames and object id's returned for the queries, total time taken by a query divided into retrieval time and evaluation time to find the correct frame by matching against the ground truth.

\begin{table*}[httb]
\small
\centering
% \caption{\tn{Missing table caption. I'd suggest adding $\uparrow$ and $\downarrow$ next to the metric names to help the reader parse the numbers a bit easier ($\uparrow$ means higher is better and so on)}}

%  $*^1$refers to the Naive baseline without D3A, $*^2$is D3A + synthetic Perfect embedding generator, $*^3$is D3A + Detectron (real world detector) with a color histogram as the embedding generator
\begin{tabular}{@{\makebox[2em][r]{\textcolor{airforceblue}\rownumber\space\space\space}} l p{1.2cm} p{1.5cm} | p{0.9cm} p{1.8cm} p{1.9cm} l p{1.9cm} p{2cm}}
\toprule
\multicolumn{1}{@{\makebox[3em][r]{}}} & \textbf{Type} & \textbf{Query Precision} & \textbf{Aggregation Method}  & \textbf{Miss Rate}$\downarrow$  & \textbf{Mean \newline Number of Frames} $\downarrow$  & \textbf{Mean  \newline Number of Object IDs} $\downarrow$  & \textbf{MRR @50} $\uparrow$ & \textbf{Mean  \newline Retrieval Time (ms)} $\downarrow$ & \textbf{Total  \newline Evaluation Time (ms) $\downarrow$}\\
\midrule
% \multirow{3}{*}{Q1} & perfect & Perfect & $0.00$ & $1.08$ & $1.00$ & $1.00$ & $0.03$ & $8.38$\\
% & perfect & Category & $0.24$ & $5.88$ & $3.75$ & $0.35$ & $0.89$ & $104.82$\\
% & perfect & Any object & $0.88$ & $38.00$ & $33.00$ & $0.07$ & $1.04$ & $2,031.02$\\
 \multirow{9}{*}{Q1}   & \multirow{3}{*}{Perfect}  & Naive                & 0.00      & 1.00        & 1.00                      & 1.00   & 0.02                    & 10.30                     \\
  &  & D3A+P          & 0.00      & 1.00        & 1.00                      & 1.00   & 0.02                    & 3.93                         \\
  \rowcolor{LightCyan} 
  &  & D3A+D                & 0.00      & 10.50       & 1.00                      & 1.00   & 0.02                    & 6.63                     \\
  \cline{2-9}

 & \multirow{3}{*}{Category}   &  Naive           & 0.33      & 792.73      & 339.00                    & 0.19   & 22.24                    & 2,567.62                   \\
&  &   D3A+P           & 0.20      & 6.29        & 4.50                      & 0.33   & 0.96                    & 101.31                   \\
\rowcolor{LightCyan} 
  &  &    D3A+D           & 0.20      & 12.43       & 1.50                      & 0.29   & 0.94                   & 104.95                   \\
 \cline{2-9}

 &  \multirow{3}{*}{Any object}  &   Naive        & 0.93      & 2,558.00    & 2,558.00                   & 0.01   & 119.74                  & 926.67                    \\
   
&  &  D3A+P         & 0.20      & 38.00       & 33.00                     & 0.24   & 0.82                    & 116.15                     \\
    \rowcolor{LightCyan} 
&   &    D3A+D          & 0.13      & 75.00       & 25.00                     & 0.20   & 1.12                    & 89.19                    \\

\hline
\multirow{9}{*}{Q2}  & \multirow{3}{*}{Perfect}      &  Naive          & 0.00      & 1.00        & 1.00                      & 1.00   & 0.02                    & 2.55                  \\

   &  & D3A+P            & 0.00      & 1.00        & 1.00                      & 1.00   & 0.02                    & 6.94                          \\
    \rowcolor{LightCyan} 
  &   & D3A+D                & 0.00      & 10.20       & 1.00                      & 1.00   & 0.03                    & 23.23             \\
   
   \cline{2-9}
& \multirow{3}{*}{Category}     &  Naive         & 0.88      & 861.00      & 861.00                    & 0.04   & 25.99                   & 2,108.60                   \\
   
   &     &   D3A+P         & 0.31      & 7.50        & 7.00                      & 0.31   & 1.63                    & 55.81                   \\
      
\rowcolor{LightCyan} 
&    &    D3A+D            & 0.38      & 18.00       & 3.50                      & 0.18   & 1.54                    & 73.26                \\
   
     \cline{2-9}
 & \multirow{3}{*}{Any object}  &  Naive         & 0.94      & 2,558.00    & 2,558.00                   & 0.01   & 117.96                 & 996.27                  \\
   
  &   &  D3A+P       & 0.25      & 38.00       & 33.00                     & 0.13   & 0.85                    & 173.65                \\
\rowcolor{LightCyan} 
  &   &     D3A+D        & 0.38      & 75.00       & 25.00                     & 0.10   & 1.14                    & 153.19                     \\

\hline

 \multirow{9}{*}{Q3}  & \multirow{3}{*}{Perfect}   & Naive          & 0.00      & 1.00        & 1.00                      & 1.00   & 0.02                   & 5.05                     \\
   &    & D3A+P            & 0.00      & 1.00        & 1.00                      & 1.00   & 0.02                    & 4.15                   \\
\rowcolor{LightCyan} 
    &      & D3A+D              & 0.00      & 5.50        & 1.00                      & 1.00   & 0.02                    & 3.53                     \\
      \cline{2-9}

& \multirow{3}{*}{Category}   &  Naive       & 0.67      & 96.00       & 96.00                     & 0.25   & 336.60                & 110.58                   \\
   
 &  &   D3A+P         & 0.27      & 1.50        & 1.50                      & 0.36   & 1.60                    & 9.20                   \\
 \rowcolor{LightCyan} 
&   &     D3A+D         & 0.47      & 1.00        & 1.00                      & 0.33 & 1.35                & 5.68                \\
         
\cline{2-9}         
 &\multirow{3}{*}{Any object}  &  Naive       & 0.86      & 684.00      & 684.00                    & 0.08   & 2,723.35                & 160.27                                    \\
   
&  & D3A+P          & 0.14      & 14.00       & 14.00                     & 0.21   & 13.25                   & 31.45                      \\

\rowcolor{LightCyan} 

  &     &    D3A+D      & 0.07      & 12.00       & 12.00                     & 0.26   & 11.51                   & 28.93                   \\
% \cline{2-9}

\bottomrule
\end{tabular}
\vspace{-2mm}
\caption{\small Exp 2 - Retrieval Performance for the three types of queries. The highlighted row corresponding to D3A+D shows performance of our deployed D3A with real detector. D3A+D's performance is better than the Naive Baseline Aggregation method and sightly worse than D3A+P (D3A's implementation with a perfect object detector) in all the query types. $\uparrow$ means higher is better, $\downarrow$  means lower is better.}
\vspace{-3mm}
\label{table:queryperformance}
\end{table*}

% \begin{table*}[!htbp]

\subsubsection{Exp 1 - Retrieval Performance with a Synthetic Embedding Generator with Increasing Uncertainty} \label{sect:falsepositiveexp}
The performance of D3A depends on the accuracy of the object detector and object pose estimates. In this experiment, we design a synthetic embedding generator that replaces the object detector and color histogram. Our synthetic embedding generator is parameterized with a false positive rate ($fpr$)\footnote{ \label{note2} $fpr$ denotes the probability of falsely labelling the detected instance} and false negative rate ($fnr$)\footnote{$fnr$ denotes the detector's probability of not detecting the object}. We keep the $fnr$ low since we assume that the object is not in the environment if the detector fails to detects the object. We annotate ground truth object embeddings as one-hot encodings $ohe$ of the unique instances, where each index represents the unique id of a ground truth object. We then use the $fpr$ and $fnr$ to flip the bits of the $ohe$ and return the transformed encoding as the synthetic embedding generator's output.
%\stnote{You don't need to give pseudocode for this if you need more space.} \tn{+1}
% We measure the MRR and Miss Rate with varying the $fpr$ - level of uncertainty of our Embedding Generator and compare it with the ``Naive" baseline. 
\textbf{Results}: We tested our system on Q1 type - \textit{Did you ever see an orange bowl?} for all ground truth object instances as mentioned in Table \ref{tab:example_queries}. Results are shown in Fig-\ref{fig:graph_noise}. Our method's MRR@50 decreases as $fpr$ increases. Despite this trend, D3A's MRR@50 remains considerably higher than Naive's MRR@50. This indicates that even when the object detector used in D3A is not good, the user will always find the correct object's frame earlier with D3A than with the Naive baseline. D3A's Miss Rate increases as the synthetic detector becomes noisier, but is always lower than Naive's Miss rate. 
The slight decreasing trend of  D3A's Miss Rate for $fpr >0.3$ is because with a higher probability of flipping the bits of the $ohe$, the probability of different object instances being assigned a similar embedding also increases. This causes D3A to consider a higher proportion of object instances to be similar, leading to more cluster formations (because of different locations) and storage of their respective keyframes. At the time of retrieval for a given object, this causes more keyframes to be returned one of which includes the queried object.
% The slight decreasing trend of  D3A's Miss Rate for $fpr >0.3$ is because with a higher probability of flipping the bits of the $ohe$, the probability of different object instances being assigned a similar embedding also increases. This causes D3A to consider a higher proportion of object instances to be similar even when they are not, leading to more detections being used for cluster formation, increasing the number of clusters formed and greater storage of their respective keyframes. At the time of retrieval this causes keyframes to be returned of more different objects and at least one of those frames includes the relevant object.
 \tn{i don't get how this means a lower miss rate}\iinote{improve this?}

\subsubsection{Exp 2 - Query Retrieval Performance}\label{sec:exp3}

Our second experiment compares the retrieval performance of two versions of D3A against the Naive Baseline. The first is $D3A+D$ - D3A with a real detector, Dectectron \cite{Detectron2018}, and color histogram and second is $D3A+P$ - D3A with a synthetic perfect sensor.
\iinote{two versions of D3A against the Naive Baseline. The first is $D3A+D$ - D3A with a real detector, Dectectron \cite{Detectron2018}, and color histogram and second is $D3A+P$ - D3A with a synthetic perfect sensor.}
% D3A with a real detector -- Dectectron \cite{Detectron2018} and color histogram ($D3A+D$) against D3A with a synthetic perfect sensor ($D3A+P$) and the ``Naive" baseline. 
We experimented over the 3 query types in Table \ref{tab:example_queries} and the 59 ground truth object instances annotated in the data.  The number of instances per query ranged from 15 -20 depending on the query, making the total number of queries to be 150. 
% \tn{How does the math work here? 3*59 is not 426, and idk where the 15-20 fits in..}
% For the given \iinote{70} queries, we test the retrieval performance  our \textit{deployed system} - on Kuri with real object detector. 

\begin{figure}
    \centering
    \includegraphics[width=0.85\linewidth]{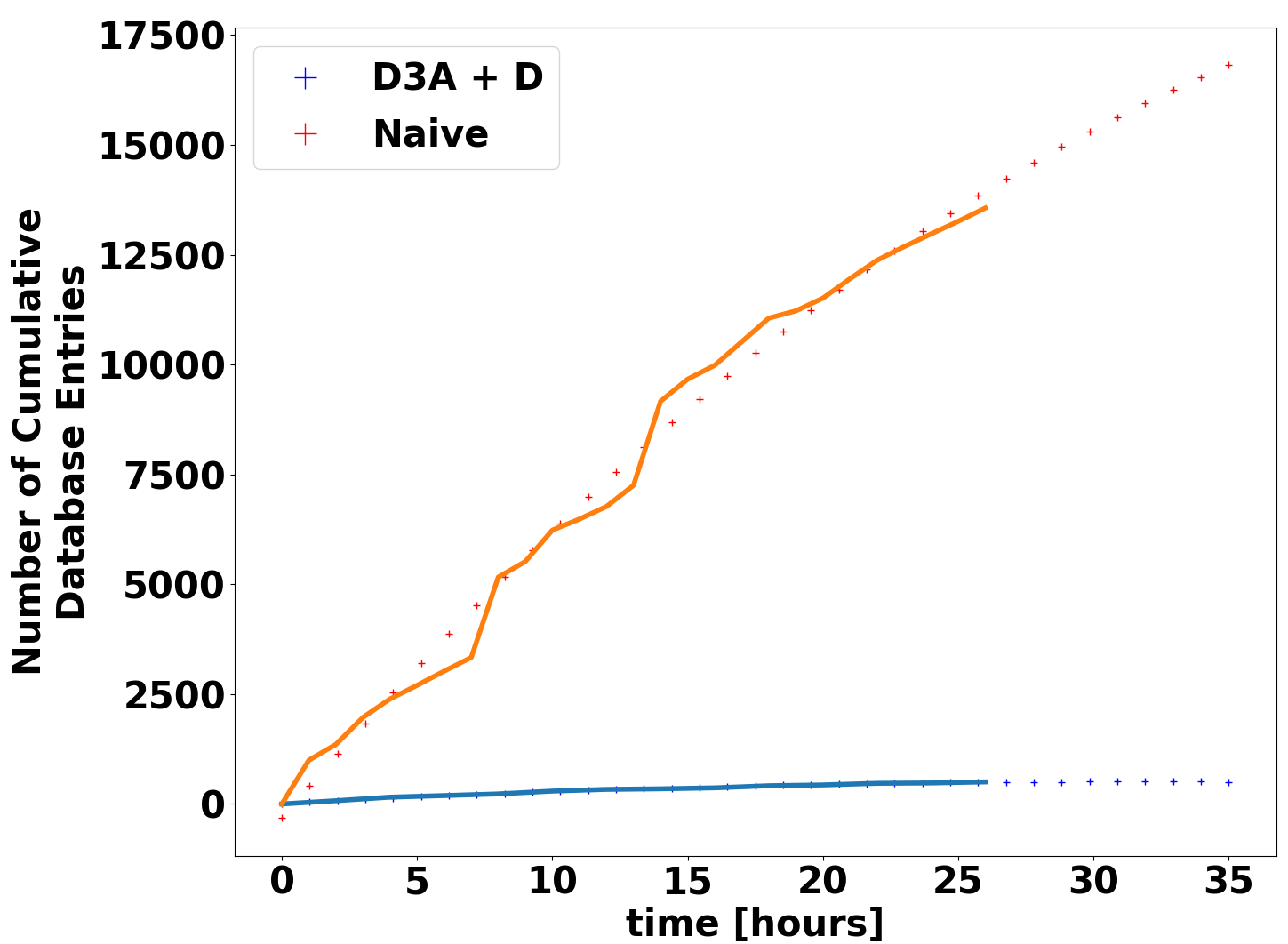}
    \caption{Exp 3 - Cumulative number of insertions to the database per hour}
    % \label{fig:a}
    % \tn{seems like there's extra white space around the figure, trim that to get back some space}
    \label{fig:graphcumulative}
    \vspace{-4mm}  
\end{figure}

\textbf{Query Matching:} Another factor affecting the retrieval performance in this experiment is the matching of the desired object in the natural language query to an $ObjectID$ in the database. To examine this axis, we discretize the query matching precision into three levels:

\begin{itemize}
    \item \textit{Perfect}: The query matching component can map an object in the language query to an exact $ObjectID$ in the database, and removes any level of uncertainty.
    \item  \textit{Category}: The matching component is noisier and is unable to uniquely identify the queried object (\textit{e.g.,} orange bowl), and instead returns a set of $ObjectID's$ for all objects from the same category (\textit{e.g.,} bowl).
    \item \textit{Any Object}: The matching component is the noisiest and cannot extract even the category-level information from the language query, and thus maps the query to the most generic set of all the $ObjectID's$ in the database.
\end{itemize}
% We report MRR for the top 50 returned frames for \textbf{U, C, A} labelled as MRR (N=50) in Table-\ref{table:results1}. 

\textbf{Results:} The results are shown in Table \ref{table:queryperformance}.
With \textit{Perfect Query Precision} for all three types of queries Q1 [\textit{rows: 1-3}], Q2 [\textit{rows: 10-12}], and Q3 [\textit{rows: 19-21}], both the mean retrieval time and total evaluation time of $D3A+D$ are less than those of the ``Naive" baseline. On average, $D3A+D$'s total (retrieval + evaluation) time is \textbf{1.96x} slower than $D3A+P$ and \textbf{1.48x} faster than the baseline. This translates to improved user experience as the users' query will be answered more quickly and as expected the miss rate is 0 because of the perfect query matching precision.

With \textit{Category Query Precision} for Q1 [\textit{rows: 4-6}], Q2 [\textit{rows: 13-15}], and Q3 [rows:\textit{22-24}], the mean number of frames returned by $D3A+D$ is \textbf{55x} smaller than that of the ``Naive" baseline, and only \textbf{5 counts}
% \tn{is this correct? you should double-check your numbers just in case} 
higher than $D3A+P$'s on average. The MRR@50 for $D3A+D$ is \textbf{0.26}, which signifies that if the queried object exists in the database, it will on average be found in the fourth frame. An empty set is returned for negative queries. While the baseline on average finds the object in the sixth frame, the probability of it not finding an object of interest in the first 50 frames (average miss rate) is \textbf{62.5\%}, which is really high compared to $D3A+D$'s \textbf{34.7\%}. On average, $D3A+D$ took \textbf{23ms} to retrieve and evaluate all the frames for all the object instances while the baseline took \textbf{500ms}. These evaluation times, however, are just for 3 hours of data. These times will increase linearly with the amount of sensor data for the ``Naive" and increase the human users' waiting time for their query to be answered.

\begin{table}[t]
\small
%\stnote{use booktabs, avoid vertical rules}
%\tn{i rounded up decimals to have max 4 decimal places.}
% \iinote{since they are so many number, I highlighted those that were a little more important and that I discuss even in the evaluation results section but maybe there is a better way to do that?} \tn{maybe add a footnote to say that ``the bolded numbers are highlighted results that are discussed in the text" or something?}  }
% \vspace{-3mm}
    \centering
    \begin{tabular}{p{3cm} l l l}
        \toprule
        {\bf{Performance Metric}} & & \bf{Naive} & \bf{D3A+D}\\
        \midrule
        Database Size (MB)  &  & $3312$  & $244$ \\
        \hline
        Total Processing Time & & $0$m $8$s &   $2$m $26$s  \\
        \hline
        Average Query Response Time (ms) & & $503.59$ & $10.63$  \\ %9.544548022
        \hline
        Total Number of Unique Object IDs &  & $\bf{2558}$ & $\bf{25}$ \\
        \hline
        \multirow{2}{3cm}{Number of Duplicates per Object}  & Mean & $ 44.10 \pm 75.66$ &  {$1.19 \pm  0.46$} \\
        & Max & $369.00$ & $3.00$ \\
        \hline
        Mean Accuracy (\%) & {} & $48.85$ & $81.98$ \\
        \hline
        \multirow{3}{3cm}{Average Number of Keyframes Returned} & 
        U & $1.00$ & $8.73$ \\
        & C & $\textbf{583.24}$ & $\textbf{10.47}$ \\
        & A & $\textbf{1933}$ & $\textbf{54.00}$ \\
        \hline
        \multirow{3}{3cm}{Mean Reciprocal Rank ($N=50$)} & 
        U & $1.00$ & $1.00$ \\ %0.9166666667
        & C & $\textbf{0.16}$	& $\textbf{0.26}$ \\ %0.4476691691
        & A & $0.03$	& $0.19$ \\ %0.3578010691 %0.4195961824 
        \bottomrule
       
    \end{tabular}
    
    \vspace{-2mm}
    \caption{System Performance (On 3 Hours of Data)}
    \footnotesize{* The bolded numbers are highlighted results discussed in the text}
    \label{table:results1}
\vspace{-3mm}
\end{table}

The worst retrieval performance is with \textit{Any Object Query Precision} for Q1 [\textit{rows: 7-9}], Q2 [\textit{rows: 16-18}], and Q3 [\textit{25-27})]. Even then, $D3A+D$ returns \textbf{54} frames on average, compared to the \textbf{30} frames returned by $D3A+P$ and \textbf{1933} frames returned by the ``Naive" baseline. This difference is also reflected in their retrieval and evaluation times. The baseline on average finds the queried object in the 25th frame ($MRR=0.04$), while $D3A+D$ and $D3A+P$ finds the object around the fifth frame. Even with the worst query precision where the queried object is mapped to all the $ObjectID's$ in the database, D3A still achieves better retrieval performance by clustering detections and thus looking through fewer instances.

%%%\stnote{This is the one where your algorithm is really really good though.  So highlight that even if the detector isn't good, this compression algorithm helps because it clusters things so you have to look through fewer results.}

% that is even when we can not figure out what the object is that we are talking about, D3A can return a small set of keyframes where the correct object of interests lies in the 5th frame.
D3A's overall accuracy is \textbf{81.98\%} with a retrieval time that is \textbf{47x} faster than the ``Naive" baseline. This can lead to improved user experience since with D3A, the user's query will be answered by searching over a smaller number of key frames (by \textbf{97\%}) than with the baseline.
% \tn{what is the 97\% trying to convey?} 

\subsubsection{Exp 3 - Compactness Comparison}
We measure the cumulative number of object id insertions in the database per hour over the complete 22 hours for both $D3A+D$ and the ``Naive" baseline. As seen in Fig-\ref{fig:graphcumulative}, the number of insertions per hour for $D3A+D$ is much less than that of the baseline and scales well as the amount of data increases, demonstrating D3A's success in aggregating partial views. D3A takes two orders of magnitude more time to process the raw data than the baseline, but outputs a representation \textbf{14.7x} more compact and efficient in answering questions, as shown in Table \ref{table:results1}. This demonstrates that D3A allows for scalable spatial-temporal representation of objects.

\subsection{Exp 4 - Comparison with the ``Non-spatial'' Baseline}
In this experiment, we visually compare the aggregated clusters formed for the unique object instances by $D3A+D$ and the ``Non-spatial" baseline. Fig-\ref{fig:clusters} shows the baseline's aggregated detections for the pink cup asked in Fig-\ref{fig:Kuri}. The baseline clusters all object instances that are similar in embedding space, irrespective of their location in the map. The images' different bounding box colors show the respective clusters they were mapped to by D3A based on both the embedding features and spatial information. The clusters formed by D3A most closely resemble ground truth. Furthermore, the ``Non-spatial'' baseline will fail to answer queries of type Q3 in Table \ref{tab:example_queries}.

\section{Discussion}

\begin{figure}
    \centering
    \includegraphics[width=0.7\linewidth]{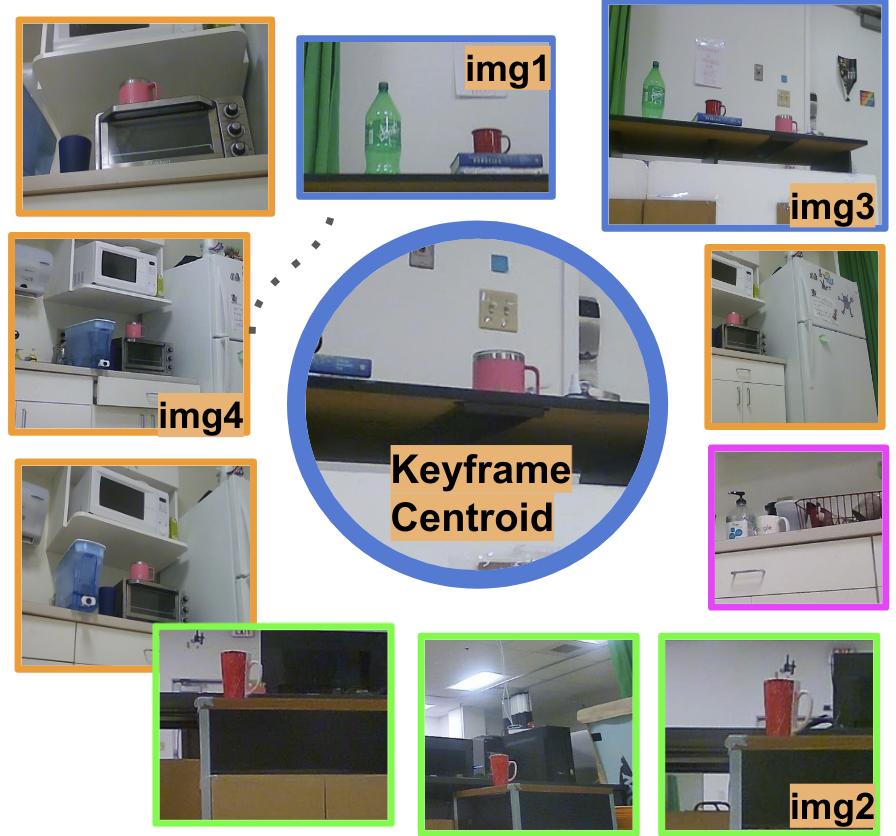}
    \caption{Exp 4 - Visualization of the keyframe centroid and the aggregated detections by the ``Non-spatial'' baseline \cite{hsieh2018focus} for the pink cup. All the square images are the detections and the different  bounding  box  colors  show  the respective  clusters  they  were  mapped  to  by  D3A  based  on both  the  embedding  feature  and  spatial  information. The dotted line connects clusters that were mapped to the same \textit{ObjectID}.}
    \label{fig:clusters}
\vspace{-4mm}
\end{figure}

% \tn{cut: Our -deployed algorithm $D3A+D$, despite performing very well, is still unable to retrieve spatio-temporal information for all the ground truth objects.}
Our deployed algorithm $D3A+D$ shows memory and speed efficiency in clustering together partial views of the same object, even when the object has moved to different locations in the environment over time. In such cases, the same object in each of its new location is assigned to a different keyframe centroid, but is associated with the same $ObjectID$. This is shown in Fig-\ref{fig:clusters} where the pink cup is moved from the table in img3 to the kitchen counter in img4 and assigned to two separate keyframe centroid by D3A, but is assigned the same $ObjectID$, symbolized by the same bounding box color. 

D3A can also disambiguate unique object instances with similar features. An example is shown in Fig-\ref{fig:clusters} where D3A assigns the two red cups in img1 and img2 to different clusters. However, our algorithm cannot perfectly retrieve all the locations a unique object was at every moment in time.
% \iinote{Add: Our deployed algorithm $D3A+D$ shows memory and speed efficiency in clustering together partial views of the same object even when moved to different locations in the environment over time. In such a case the same object in each of it's new location is assigned to a different key frame centroid but gets associated to the one object\_id. This is shown in Fig.\ref{fig:clusters} where the pink cup is moved from the table ``img3" to the kitchen counter ``img4" and is assigned two separate keyframe centroid by D3A since they are two locations but is assigned the same object\_id. D3A can also disambiguate unique objects instances even if they have similar feature. An example is shown in Fig.\ref{fig:clusters} where D3A assigns the two similar red cups in ``img1" and ``img2" different clusters. Despite such results, our algorithm is unable to perfectly retrieve all of the locations where a unique object was placed at every moment in time.} 
% spatio-temporal information for every ground truth object.}
%%%Our deployed algorithm $D3A+D$, despite performing very well, is still unable to retrieve spatio-temporal information for all the ground truth objects.
% Overall, our deployed algorithm $D3A+D$ performs very well and correctly retrieves objects most of the time. However, it is still unable to retrieve the spatial-temporal information for all ground truth object instances. 
This is because our system sometimes fails to distinguish objects when they have similar embedding features and are close together in Euclidean space (below $d_{thresh}$). In this case, D3A will assign the same $ObjectID$ and keyframe centroid to two unique object instances. 
% This will cause D3A to miss storing the spatial-temporal information of the wrongly assigned object and. 
An example of this is shown in Fig-\ref{fig:clusters} where img1 of the red cup is assigned to the cluster for the pink cup since they have similar colors and are close together in the physical world. 
This causes D3A to miss the associated information of the wrongly assigned red cup and not be able to retrieve this information in response to an object retrieval query related to the red cup.

Furthermore, if an object is only seen in one frame of a sliding window, our algorithm will characterize it as noise to account for errors in the detection. This, however, causes D3A to sometimes miss unique object instances. Overall, D3A recognized 76 unique object locations in the map, but because of the above reasons assigns them to 25 object instances instead of the ground truth 59, while considering 18 of the objects as noise. As shown in Table \ref{table:results1}, the average number of duplicated $ObjectID's$ assigned to a ground truth object instance by D3A is $1.19 \pm 0.46$, which is 36X smaller than that of ``Naive."
% \iinote{Add: In case of Naive every partial view detection of the object is treated as a unique object leading to total number of unique detections to be greater than the total number of ground truth objects.}
In the case of the ``Naive" baseline, every partial view detection of the object is treated as a unique object, leading to a total number of unique detections much greater than the number of ground truth objects. 
Fig-\ref{fig:graph_noise} and Table \ref{table:queryperformance} show that increasing the false positive rate of the visual detector and decreasing the query precision degrades object retrieval performance from the spatial-temporal representation created by D3A. Nevertheless, the deployed D3A+D algorithm on Kuri with the real-world detector performs better than the baseline in terms of memory and speed efficiency for object retrieval. D3A+D also performs better than ``Naive" despite the fact that they both suffer from drifting localization of the robot's pose over time. This is because D3A's three-tier clustering and filtering structure was able to filter out some of the localization drift as noise in the objects' sensor data.

We took the first step in aggregating the detections of objects over space and time with our proposed algorithm - D3A, and defer the reduction of duplicates by using improved deep-learned object embeddings instead of color histograms to future work. D3A can enable service robots to find misplaced objects for the elderly over long periods of time.
%The lost objects could be misplaced while remaining in robot's line of sight.
Our research also opens venues for object retrieval in other surveillance related applications. Our work has only been evaluated in an environment where the pre-built map was provided to us. It will be interesting to explore how our system can be adopted to work in unknown open domain environments. We plan to conduct a human user study in the future to investigate their experience when querying D3A and searching over the small subset of returned keyframes to find answers to their object retrieval queries.
\vspace{-3mm}
\section{Conclusion}
We present a novel algorithm for robots to efficiently answer spatial-temporal queries about objects in the environment over long periods of time. Our algorithm aggregates partial view detections of unique instances to create a compact and query-able representation of the objects. By explicitly performing detection-based three-level association to identify the keyframes for unique object instances, our algorithm significantly outperforms baselines in answering queries in terms of the retrieved frames' accuracy, mean reciprocal rank and retrieval time. A robot deployed with our algorithm was able to process 22 hours of 
% \iinote{I evaluate on 3 hours? but do make compact for 22 hours so should it be 22 hours?} \tn{i'd say 22 bc it did process and compress that much data, we just don't know how well}
sensor data and develop a compact representation of objects in its environment, which is an encouraging step towards enhancing the sensory capabilities of home-service robots that can help the elderly find their lost, forgotten objects.

\section{Acknowledgments}
This work was supported by NSF under grant award IIS-1652561 and an award from Echo Labs. The authors would like to thank Baber Khalid, Thao Nguyen, and Eric Rosen for their valuable feedback and insights on the research and Baber Khalid for being a supportive husband.
\bibliography{references}

\newpage
\section{Appendix}
% \section{Graph Appendix} \label{appendix:graph}
% \appendix
\subsection{Algorithmic Implementation} \label{sec-Algo Implementation}

We  deploy D3A  on  a  mobile  robot  Kuri \cite{heykuri_2018} that patrolled a robotics lab environment for four days. Our system performs real-time object detection with Facebook's AI Research software system - Detectron \cite{Detectron2018} to extract $c_{ij}, prob_{ij}$ for every detection. We use a color histogram \cite{color_hist} to approximate an embedding $emb_{ij}$ of each object instance detection. We select the color histogram method due to its simplicity. However, our algorithm is not constraint to using the color histogram as the instance feature extractor. More sophisticated deep learning methods that use RGB or RGB-D images \cite{bras2018unsupervised, xu2014deep} can also be used. 

In the subsection Experiment 1 of section Evaluation, we show how varying the type of object detector (changing its false positive rate) affects the results of our algorithm. Notably, our approach is also not constrained to any particular number of objects, object size or shape, or object dynamicness (\textit{i.e.,} the objects' locations can change over time). We are only constrained by the object categories the detector was trained to detect. Processing of the visual data and pose estimates was done off-board on a stand-alone machine. The PC setup included an Intel(R) Core(TM) i7-6700 CPU (3.40GHz x 8), and a GeForce GTX 1070/PCIe/SSE2 GPU running Ubuntu 16.04. D3A was run online while the PC communicated with the Kuri robot over ROS \cite{quigley2009ros}. For Tier 1-clustering of detections across adjacent frames we use density-based spatial clustering (DB-Scan) \cite{khan2014dbscan}, a method known to be good at handling outliers/noise within the dataset. While other clustering methods can also be used, we defer the analysis of their impact on D3A's performance for future work.

All the aggregation operations $\odot$, $\oplus$ and $\otimes$ that we applied are the weighted average functions. These aggregation functions can also be different. Weighted average was selected because of its simplicity and reasonable empirical performance for object retrieval on our smaller dataset of $<3$ hours of data. More sophisticated methods can also be used in place of the weighted average. However, as shown in section -- Experiment Designs of Evaluation,  
% -\ref{sec:exp-des},
with just the weighted average, D3A still performs better than the baseline.

We organized our aggregated spatial-temporal representation $R$ over two collections in \textit{MongoDB} \cite{banker2011mongodb}, a database that supports large volumes of both data and traffic. The two document Mongo-DB collections \textbf{Object Identification Collection} ($OIc$), and \textbf{Spatial-temporal Collection} ($STc$) are as described in Section -- Tier 3- Filtering across database entries of Approach.
% Sec-\ref{subsubsection:p3}. 
$OIc$ in our implementation, stores the object information using the associate embedding of the cluster, the category assigned to the cluster and a weight which encodes the frequency of detections for this object and is used in the aggregation operation between $record$ in $OIc$ and $lrv\_entry$ evicted by the $STM$.
% \tn{the previous sentence doesn't make sense? Do you mean something like ``The two collections, $OIc$ and $STc$ are as described in section..."?}
For natural language processing of the queries, we annotate the queries (use the template-based method) to convert them into the logical form shown in Table \ref{tab:example_queries}. These logical forms are then used to query the collections for relevant spatial-temporal information. In the subsection Experiment 2 of section Evaluation we show how the performance of D3A changes with varying the uncertainty in the processing of the query but still it is able to perform better than the baselines. 

% \iinote{We organize our aggregated spatial-temporal representation over two collections in \textit{MongoDB} \cite{banker2011mongodb}, a database that supports large volumes of both data and traffic. Our first collection in MongoDB is the \textbf{Object Identification Collection} ($OIc$) that stores object attributes of aggregated clusters such as the category, embedding and weight. We also have a separate \textbf{Spatial-temporal Collection} ($STc$) that stores documents containing a geo-spatial index for location, time intervals, the keyframe, and the detection probability and bounding box of the object in the keyframe.}
% \iinote{We annotate the queries to convert to the database query representation.}

% \section{Additional Results} 
% \end{appendix}

% \appendix 

\end{document}